\pdfoutput=1
\documentclass[11pt]{article}

\usepackage{naacl2021}

\usepackage{times}
\usepackage{latexsym}

\usepackage[T1]{fontenc}
\usepackage[utf8]{inputenc}

\usepackage{microtype}

\usepackage{booktabs} 
\usepackage{multirow}
\usepackage{url}
\usepackage{amsmath}
\usepackage{bm}
\usepackage{amsfonts}
\usepackage{graphicx}
\usepackage{diagbox}
\usepackage{xspace}
\usepackage{courier}
\usepackage{mathrsfs}
\usepackage{appendix}
\usepackage{float}
\usepackage{framed} 
\usepackage{color}
\usepackage{xspace}
\usepackage [switch]{lineno}

\newcommand\DTBERT{TR-BERT\xspace}
\newcommand\BASESIZE{$_{\small \textsc{base}}$\xspace}

\newcommand\BERTL{BERT$_{\text{L}}$\xspace}
\newcommand\DTBERTL{TR-BERT$_{\text{L}}$\xspace}
\newcommand\DTBERTD{TR-BERT$_{12}$\xspace}
\newcommand\DTBERTS{TR-BERT$_6$\xspace}

\title{\DTBERT: Dynamic Token Reduction for Accelerating BERT Inference}

\author{Deming Ye$^{1,2}$, Yankai Lin$^{4}$, Yufei Huang$^{1,2}$,  Maosong Sun$^{1,3}$\thanks{ \ \ Corresponding author: M. Sun (sms@tsinghua.edu.cn)}\\
$^1$Department of Computer Science and Technology, Tsinghua University, Beijing, China\\
Institute for Artificial Intelligence, Tsinghua University, Beijing, China\\
Beijing National Research Center for Information Science and Technology\\
$^2$State Key Lab on Intelligent Technology and Systems, Tsinghua University, Beijing, China\\
$^3$Beijing Academy of Artificial Intelligence \\
$^{4}$Pattern Recognition Center, WeChat AI, Tencent Inc.\\
\texttt{ydm18@mails.tsinghua.edu.cn}
}

\begin{document}
\maketitle
\begin{abstract}
Existing pre-trained language models (PLMs) are often computationally expensive in inference, making them impractical in various resource-limited real-world applications. To address this issue, we propose a dynamic token reduction approach to accelerate PLMs' inference, named \DTBERT,  which could flexibly adapt the layer number of each token in inference to avoid redundant calculation. Specially, \DTBERT formulates the token reduction process as a multi-step token selection problem and automatically learns the selection strategy via reinforcement learning. The experimental results on several downstream NLP tasks show that \DTBERT is able to speed up BERT by $2$-$5$ times to satisfy various performance demands.
Moreover, \DTBERT can also achieve better performance with less computation in a suite of long-text tasks since its token-level layer number adaption greatly accelerates the self-attention operation in PLMs. The source code and experiment details of this paper can be obtained from \url{https://github.com/thunlp/TR-BERT}.
%We will release the source codes and datasets for further research explorations.

%accelerate inference by $1$-$5$ times without significant performance degradation compared to BERT. 

%through the token-level layer number adaption, \DTBERT could process long sequences more effectively as it greatly accelerates the self-attention operation in PLMs. The results show that, compared to BERT, \DTBERT can also achieve better performance with less computation in a suite of  long-text tasks. 
\end{abstract}

\section{Introduction}

Large-scale pre-trained language models (PLMs) such as BERT~\cite{BERT}, XLNet~\cite{XLNET} and RoBERTa~\cite{RoBERTa} have shown great competence in learning contextual representation of text from large-scale corpora. With appropriate fine-tuning on labeled data, PLMs have achieved promising results on various  NLP applications, such as natural language inference~\cite{nlibert}, text classification~\cite{sentimentbert}  and question answering~\cite{qabert}. 
% relation extraction~\cite{peters-etal-2019-knowledge},

Along with the significant performance improvements, PLMs usually have substantial computational cost and high inference latency, which presents challenges to their practicalities in resource-limited real-world applications, such as real-time applications   and hardware-constrained mobile applications. Even worse, these drawbacks become more severe in long-text scenarios because self-attention operation in PLMs scales quadratically with the sequence length. Therefore, researchers have made intensive efforts in  PLM's inference acceleration recently. The mainstream approach is to reduce the layer number of PLMs such as knowledge distillation models~\cite{DistilBERT,BERT-PKD}, and adaptive inference models~\cite{deebert, FastBERT}. Such layer-wise pruning reduces a tremendous amount of computation, but it sacrifices the models' capability in complex reasoning. Previous works~\cite{DistilBERT,BERT-PKD} have found that the shallow model usually performs much worse on the relatively complicated question answering tasks than text classification tasks. It is straightforward that pruning the entire layer of PLMs may not be an optimal solution in all scenarios. 

% to accelerate PLMs' inference, aiming 
In this paper, we introduce a dynamic token reduction method \DTBERT to find out the well-encoded tokens in the layer-by-layer inference process, and save their computation in subsequent layers. The idea is inspired by recent findings that PLMs capture different information of words in different layers (e.g., BERT focuses on the word order information~\cite{Bertlow} in the bottom layers, obtains the syntactic information~\cite{Bertmiddle} in the middle layers, and computes the task-specific information in the top layers~\cite{BERTology}). Hence, we could adapt different tokens to different layers according to their specific roles in the context. %rather than transit all tokens through the whole PLM. %It is more flexible and elaborate, leading to a more elaborate model pruning and acceleration. 

%different layers in BERT capture different information of words. For example, BERT exacts the word order information~\cite{Bertlow} in the bottom layers, obtains the syntactic information~\cite{Bertmiddle} in the middle layers, and computes the task-specific information in the top layers~\cite{BERTology}. 
%These findings inspire that we could adapt the layer number for different tokens according to their specific roles in the context, leading to a more elaborate model pruning and acceleration. 

\begin{figure*}[t!]
    \centering
    \includegraphics[width=0.92\textwidth]{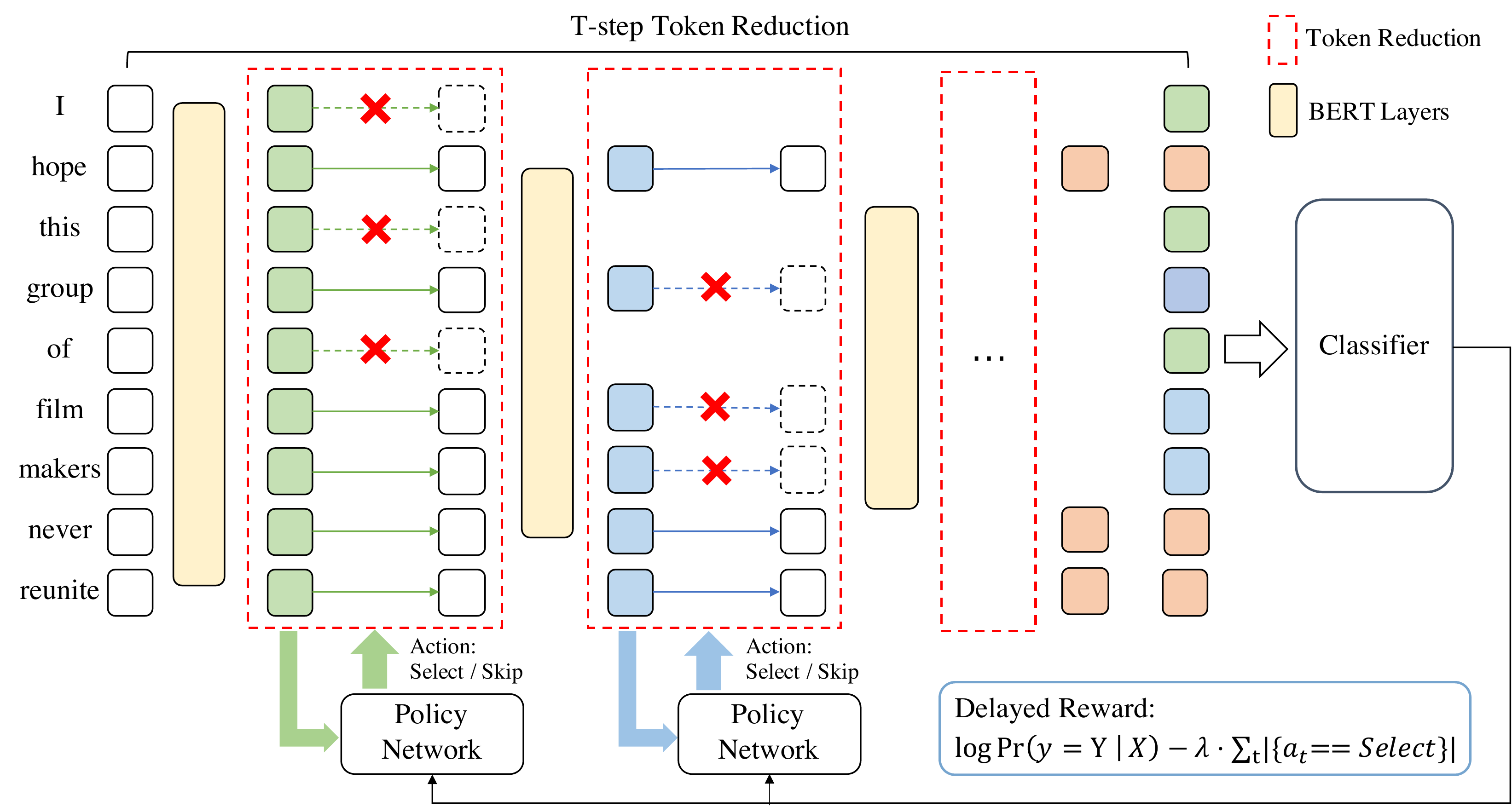}
    \caption{An illustration of the \DTBERT. \DTBERT gradually compresses the text sequence as the layer gets deeper. In RL training, we sample actions of \emph{Select} or \emph{Skip} for each token via the policy network. }% To select the important tokens, \DTBERT employs reinforcement learning to learn the selection strategy. In a one-step reduction, we sample actions of \emph{Select} or \emph{Skip} for each token via policy network. The reward is weighted by the prediction probability of pruned networks and the number of tokens reserved to balance the accuracy and speed.}
    \label{fig:model}
\vspace{-1em}
\end{figure*}
As shown in Figure~\ref{fig:model}, \DTBERT formulates the token reduction process as a multi-step selection problem. Specially, for each selection phase, \DTBERT finds out the words that require high-level semantic representations, and then selects them to higher layers. The main challenge in \DTBERT is how to determine each token's importance for text understanding in the token selection. It is highly task-dependent and requires to consider the correlation and redundancy among various tokens. \DTBERT employs the reinforcement learning (RL) method to learn the dynamic token selection strategy automatically. After the token reduction, the RL reward involves the confidence of the classifier's prediction based on the pruned network to reflect the quality of token selection. Moreover, we also add a penalty term about  the number of selected tokens  to the reward, by adjusting which, \DTBERT can utilize the different pruning intensities in response to various performance requirements. In \DTBERT, by selecting a few important tokens to go through the entire pipeline, the inference speed turns much faster and no longer grows quadratically with the sequence length.

%To find an effective token selection scheme faces two main challenges: (1) how to determine the importance of each token, which is highly task-dependent and difficult to measure; (2) how to consider the token correlation in the selection and the redundancy between the similar important tokens.  

%To address this issue, \DTBERT employs the reinforcement learning (RL) method to automatically learn the dynamic selection strategy. It computes the RL reward by two parts: (1) The confidence of classifier's prediction based on the  pruned network to reflect the token reduction quality; (2) The length punishment of the remaining tokens, by adjusting which, \DTBERT can utilize the different pruning intensity in response to various performance requirements.

We conduct experiments on eleven NLP benchmarks. Experimental results show that 
%\DTBERT accelerates the BERT inference by $1$-$5$ times without significant performance drop. 
\DTBERT can accelerate BERT inference by $2$-$5$ times to meet various performance demands, and  significantly outperform previous baseline methods on question answering tasks. It verifies the effectiveness of the dynamic token reduction strategy. %Incorporated with the layer-wise methods, \DTBERT could make a more reasonable model structure for different tasks. Besides, 
Moreover, benefiting from the long-distance token interaction, \DTBERT with $1{,}024$ input length reaches higher performance with less inference time compared to the vanilla BERT in a suite of long-text tasks.

\section{Background and Pilot Analysis}

To investigate the potential impact of the token reduction in PLMs, we first introduce the Transformer architecture. After that, we conduct pilot experiments as well as empirical analyses for the lower and upper bound of the token reduction in this section.

The Transformer architecture~\cite{Transformer}  has been widely adopted by the pre-trained language models (PLMs) for inheriting its high capacity. Basically, each Transformer layer wraps a Self-Attention module (Self-ATT) and a Feed-Forward-Network module (FFN) by the residual connection and layer normalization.  Formally, given a sequence of  $n$ words, the hidden state of the $i$-th layer, $\bm{H}_{i} = (h_1, h_2, \ldots, h_n)$,  is computed from  the previous layer state: 
\begin{align}
    \bm{M}_{i-1} &= \text{LN}(\bm{H}_{i-1} + \text{Self-ATT}(\bm{H}_{i-1}) ), \nonumber\\
    \bm{H}_{i} &= \text{LN}(\bm{M}_{i-1} + \text{FFN}(\bm{M}_{i-1}) ), 
\end{align}
where $i\in [1, L]$, $L$ is the number of stacked Transformer layers, LN denotes the LayerNorm layer. %After all layers' operations, the output states of the last layer, $H_L$, are fed to a specific output network to complete different downstream tasks.
For each Transformer layer, the complexity of the Self-Attention module scales quadratically with the sequence length. Therefore, the speed of Transformer architecture will decline heavily  when the sequences become longer.

\begin{figure}[!t]
\centering
\begin{minipage}{0.49\linewidth}
  \centerline{\includegraphics[width= 1.15\textwidth]{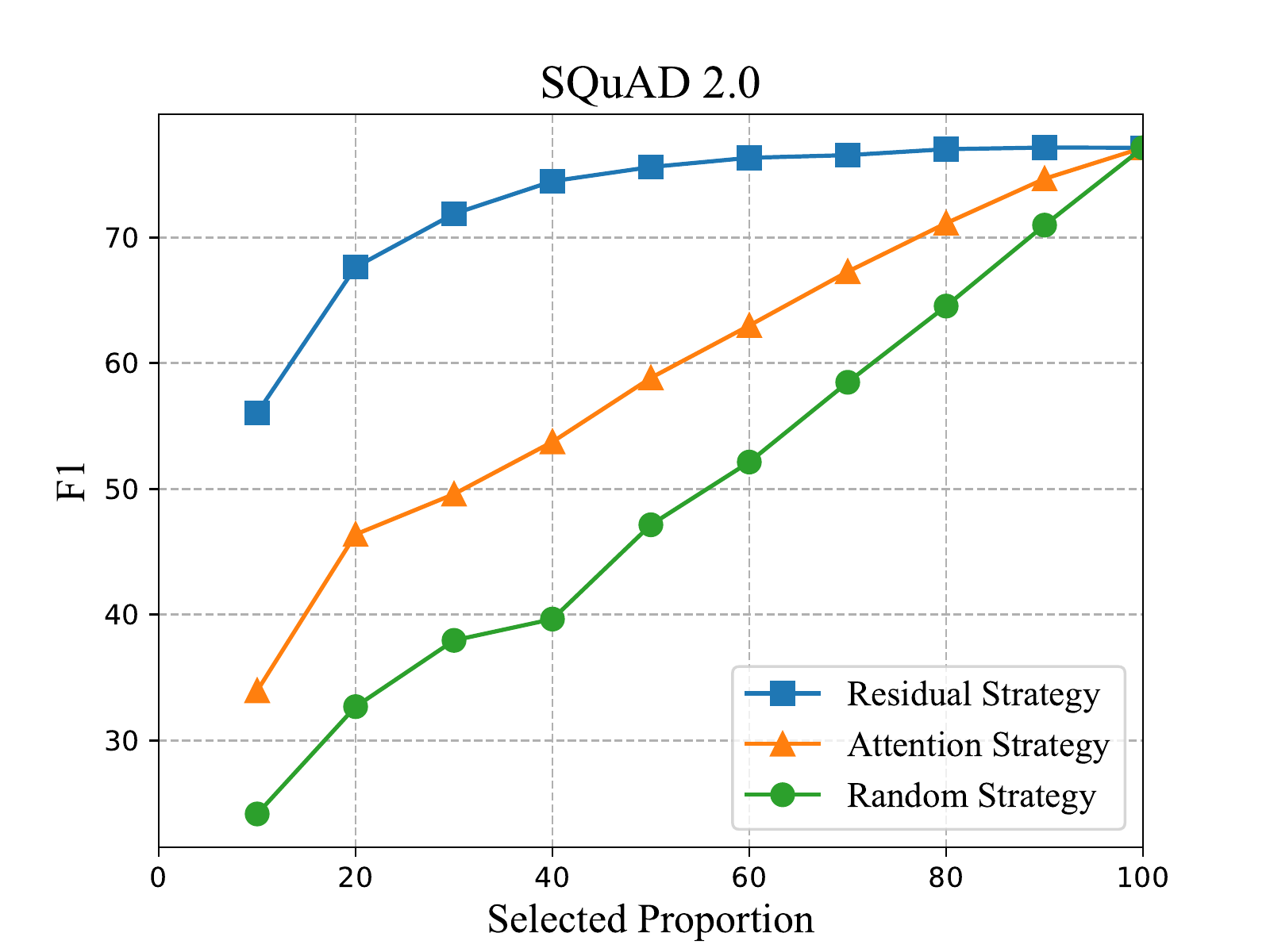}}
\end{minipage}
\hfill
\begin{minipage}{0.49\linewidth}
  \centerline{\includegraphics[width= 1.15\textwidth]{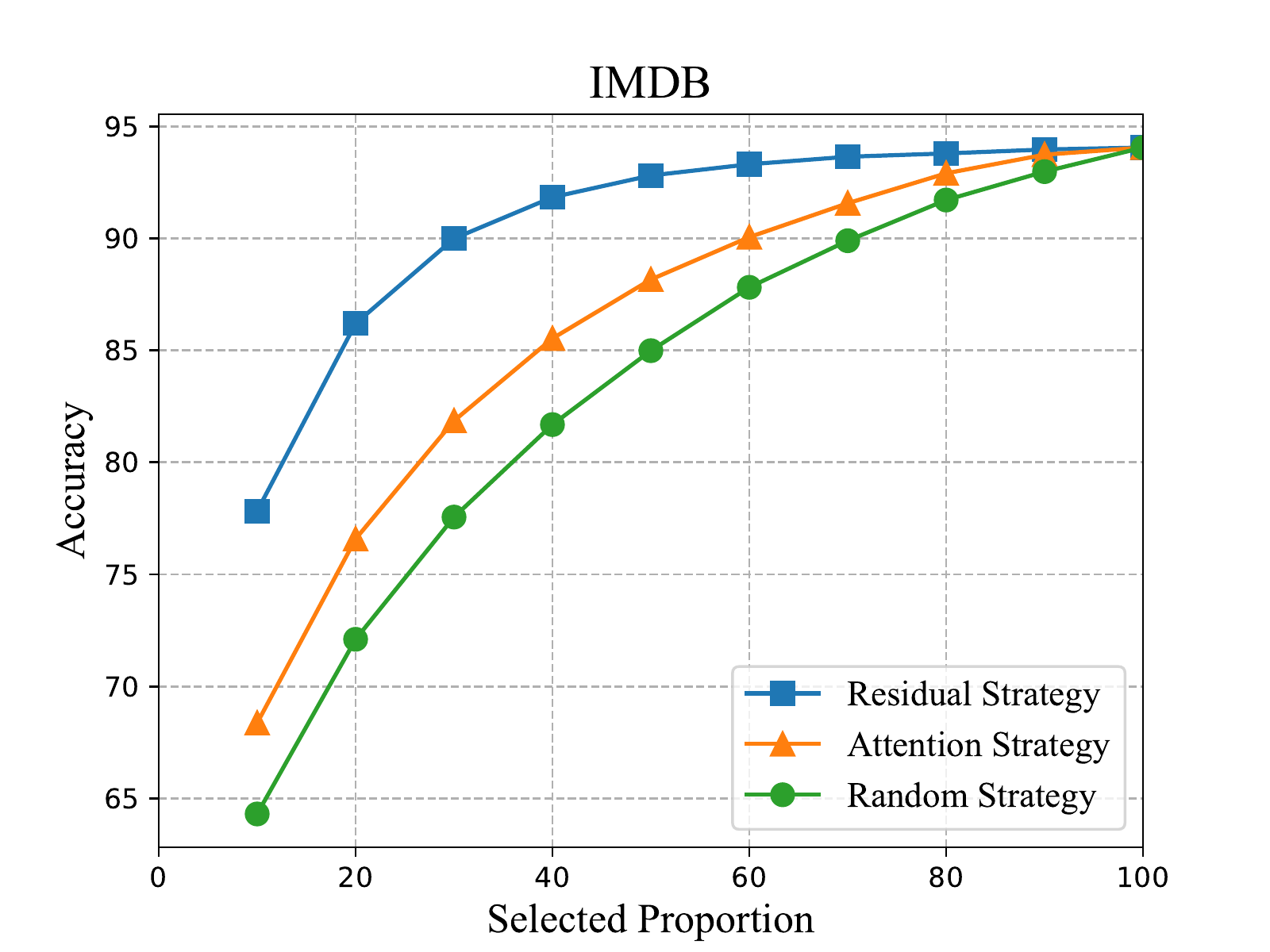}}
\end{minipage}
\caption{Performance under different selected proportion of tokens and different heuristic strategies.}
\label{fig:strategy}
% \vspace{-1em}
\end{figure}

Previous findings~\citep{BERTology} reveal that some words, such as function words, do not require high-layer modeling, since they store little information and have been well handled by PLMs in bottom layers. Hence, selecting only the important words for high-layer computation  may be a possible way to accelerate the PLMs' inference.

To verify this assumption, we conduct a theoretical token elimination experiment in question answering (on  SQuAD\,2.0~\cite{SQuAD2}) and text classification (on IMDB~\cite{IMDB}). We use the full-layer representations for the selected tokens and  the early-layer representation of the deleted tokens for the prediction. To be specific, we  eliminate tokens immediately after the $l$=$4th$ layer and adopt the following three strategies to select the retained tokens:

\textbf{Random Strategy (Lower Bound)}  selects tokens randomly, assuming that all tokens are equivalent for understanding.

%text information is uniformly distributed. 

\textbf{Residual Strategy (Upper Bound)} directly utilizes the model prediction of the original model to guide the token selection. Specially, we define a token's importance according to the influence on the model prediction when it's not selected. When substituting the $r$-th layer representation $\bm{H}_{r}$ with the $l$-th layer representation $\bm{H}_{l}$ ($r > l$) ,   we define the approximate variation to  model loss as the token importance: $\bm{I}  =  \frac{\partial {loss}}{\partial \bm{H}_{r}} (\bm{H}_{r} - \bm{H}_{l})$. 
% Here, we select a different proportion of tokens after $l=4$ layer and set $r=9$  for the Residual Strategy. 
Here, we set $r=9$ since other values get a little worse results. 
Note that we could not obtain the model loss in the prediction stage. Hence, the Residual Strategy could be viewed as an upper bound of token selection to some extent when we ignore  the correlation and redundancy among the selected tokens. 

\textbf{Attention Strategy}  is adopted by PoWER-BERT~\cite{PoWER-BERT} and L-Adaptive~\cite{lenthadaptive}. It accumulates the attention values from other tokens to a given token.  It selects the tokens receiving the greatest attentions, considering them responsible for retaining and disseminating  the primary information of the context.% to understand the text.
%Attention Strategy considers those tokens, which receiving the greatest attentions, are responsible for retaining and disseminating the primary information. Thus it selects those tokens for better text understanding.

As shown in Figure~\ref{fig:strategy}, both Attention Strategy and Residual Strategy achieve considerable results, which demonstrates that to select important tokens is feasible for accelerating the inference of PLMs. Besides, the Residual Strategy outperforms the Attention strategies by a margin, especially at the low token remaining proportion ($+31.8\%$ F1 on SQuAD 2.0 and $+9.5\%$ accuracy on IMDB when selecting $10\%$ tokens). It suggests that the accumulated attention values still cannot well reflect tokens' importance in text understanding, which requires further explorations.

\section{Methodology}

In this section, we present \DTBERT, which adopts a cascade token reduction to prune the BERT model at token-level granularity dynamically. In a one-step token reduction process, \DTBERT estimates the importance of each token, reserves the important ones, and delivers them to the higher layer. 
To better select important tokens for text understanding while satisfying various acceleration requirements, we employ the reinforcement learning (RL) method to automatically learn a dynamic token selection strategy. 
%which can also take the correlation of selected tokens into account.
%We employ the reinforcement learning (RL) method to automatically learn the dynamic token selection strategy to better measure the token importance in text understanding. 
%Benefiting from the compression of sequence length and the effective token selection, \DTBERT offers an inspiriting trade-off between accuracy and inference time.

\subsection{Model Architecture}

Figure~\ref{fig:model} shows the model architecture of \DTBERT. To inherit the high capacity from the PLMs, \DTBERT keeps the same architecture as BERT. Differently, as the layer gets deeper, \DTBERT gradually shortens the  sequence length via token reduction modules, aiming to reduce the computational redundancy of unimportant tokens.% for understanding the whole text. 

The token reduction modules are required to measure the importance of tokens and offer an integral selection scheme. Due to the lack of direct supervision, we employ the policy network for training the module, which adopts a stochastic policy and uses a delayed reward to guide the policy learning. In one-step reduction, we perform action sampling for the current sequence. The selected tokens are conveyed to the next Transformer layer for further computation. In contrast, the unselected tokens are terminated with their representation remaining unchanged. After all the actions are decided, we fetch each token's representation from the layer where it terminated, and compute the golden label's likelihood as a reward. To be specific, we introduce state, action, reward, and objective function as follows: 

\textbf{State} State $s_t$ consists of the token representations inherited from the previous layer before the $t$-th token reduction layer.

\textbf{Action} We adopt two alternative actions for each token, \{\emph{Select}, \emph{Skip}\}, where the token can be selected for further computation or be skipped to the final layer. We implement the policy network as a two-layer feed-forward network with GeLU activation~\cite{GeLU}: 
\begin{equation}
\pi(a_t|s_t; \bm{\theta} ) = \sigma (\bm{W}_2 (\text{GeLU}(\bm{W}_1 \bm{H}_{s_t} + {b}_1) ) +{b}_2),
\end{equation}
where  $a_t$ denotes the action at state $s_t$ for sequence representation $\bm{H}_{s_t}=\{{h}_{1}, h_2, ..., {h}_{n}\}$ at $t$-th reduction,  $\bm{\theta}=\{\bm{W}_1, \bm{W}_2, {b}_1, {b}_2\}$ are trainable parameters, and $\sigma(.)$ is sigmoid activation function. For the selected token set $\{ t_1, t_2, ..., t_{n^{*}}\}$, where $n^{*}  \leq n$, we conduct a Transformer layer operation on their corresponding representations:
\begin{equation}
    \mathbf{H}' = \text{Transformer}([{h}_{t_1}, {h}_{t_2}, \ldots , {h}_{t_{n^{*}}}] ).
\end{equation}
For the selected tokens, their representation $\mathbf{H}'$ is conveyed to the next layer for further feature extraction and information aggregation. For the other skipped tokens, their representations in the current layer are regarded as their final representations.

\textbf{Reward} Aiming to select significant tokens for making a precise decision in the prediction layer, we adopt the likelihood of predicting the golden label as a reward.
For example, when classifying the input sequence $\bm{X}$, 
we use the models' predicting  probability of the ground-truth label $\bm{Y}$ to reflect the quality of the token selection. In addition, to encourage the model to delete more redundant tokens for accelerating, we include an additional punitive term by counting the number of selected tokens. Hence, the overall reward $R$ is defined as:
\begin{equation}
    R = \log \Pr (y=\bm{Y}|\bm{X}) -  {\lambda}\cdot\sum_{t}|{\{a_t=\text{\emph{Select}}}\}|,
\end{equation}
where $\sum_{t}|\{{a_t=\text{\emph{Select}}}\}|$ denotes the total number of the selected tokens in all token reduction modules, and $\lambda$ is a harmonic coefficient to balance two reward terms. 

\textbf{Objective Function} We optimize the policy network to maximize the expected reward. Formally, our objective function is defined as:
\begin{eqnarray}
      J(\bm{\theta}) &=& \mathbb{E}_{(s_t,a_t) \sim \pi(a_t|s_t; \theta)} r[(s_1,a_1)...(s_T,a_T)]       \nonumber\\
           &=& \sum_{(s_1,a_1)...(s_T,a_T)} \prod_t \pi_{\theta} (a_t|s_t) \cdot R, 
\end{eqnarray}
where $T$ is the number of states. According to the REINFORCE algorithm~\cite{RL1} and policy gradient method~\cite{RL2}, we update  network with the policy gradient as below:
\begin{equation}
    \nabla_{\theta} J(\theta) = \sum_{t=1}^T R \cdot \nabla_{\theta} \log \pi_{\theta} (a_t|s_t).
\label{rleq}
\end{equation}

% \subsection{Knowledge Distillation}

% We adopt Knowledge Distillation (KD)~\cite{KD} to transfer knowledge from the intact teacher model to the pruned student model. KD can be modeled as minimizing the 
% KL-Divergence between teacher soft-label $p_t(i)$ and student prediction $p_s(i)$:
% \begin{equation}
%     L_{KD} = -\sum_{i=1}^{N} p_t(i) * \log\frac{p_s(i)}{p_t(i)},
% \end{equation}
% where $N$ is the number of labels.

\subsection{Model Training}

Our policy network is integrated into the original Transformer network, and we train both of them simultaneously. The entire training process involves three steps:

(1) Fine-tune the PLM model for downstream tasks with the task-specific objective;

(2) Freeze all the parameters except that of the policy network, conduct reinforcement learning (RL), and update the policy network to learn token reduction strategy;

(3) Unfreeze all parameters and train the entire network with the task-specific objective and RL objective simultaneously. 

%\end{enumerate} %(1) (2)   (3) 

Due to the large searching space, RL learning is difficult to converge. We adopt imitation learning~\cite{ImitationLearning} for warming up the training of the policy network. To be specific,  in the RL training,  we sample several action sequences via the policy network to compute rewards. And we guide the optimization direction by providing heuristic action sequences sampled by the Residual Strategy during the early training period, which could roughly select the most important tokens. The heuristic action sequence is defined as selecting the top $K$ important tokens and skipping the others, where $K$ is defined as the expected selected number of the current policy network. In our preliminary experiment, both the heuristic action sequence and expected selected number mechanism are beneficial to the stable training. 

To further improve the performance of our pruned model, we also adopt Knowledge Distillation (KD)~\cite{KD} to transfer knowledge from the intact original fine-tuned model.

\subsection{Complexity Analysis}

For a Transformer layer with a hidden size of $d$ and an input sequence of $n$ tokens, the Self-Attention module consumes $O(n^2d)$  time and memory complexity while the Feed-Forward Network takes $O(nd^2)$. That is, our token reduction gains near-linear speedup when $n$ is relatively smaller than $d$. Therefore, when the input sequence gets longer, such as up to $1{,}024$ tokens, our method can enjoy a more effective speedup.% In contrast, the two-layer policy network of \DTBERT makes the input dimension vary as $768$->$32$->$1$. The increased inference cost for our method is very tiny compared to the saved computation.

In the RL training, we compute loss on the pruned model, so the acceleration is still valid for this stage. Since we focus on accelerating BERT inference, we consider the extra training consumption on the pruned model is acceptable.

%For example, if we sample $8$ action sequences per step on a $4$x speed-up model, the training cost is about twice as much as the original fine-tuning. 

\section{Experiment}

In this section, we first introduce the baseline models and the evaluation datasets. After that, we verify the effectiveness of \DTBERT  on eleven NLP benchmarks. Finally, we conduct a detailed analysis and case study on \DTBERT to investigate the selected tokens' characteristics.

% counterparts

\subsection{Baselines}

We adopt two pre-trained models and three pruned networks as our baselines for comparison:

\textbf{BERT}~\cite{BERT} is a Transformer-based pre-trained model.  We use the BERT\BASESIZE model\footnote{https://github.com/google-research/bert}, which consists of $12$ Transformer layers and  supports a maximum sequence length of $512$.
% on Wikipedia\footnote{https://en.wikipedia.org/} and BookCorpus~\cite{Bookcorpus}
%  released by Google\footnote{https://github.com/google-research/bert}
% with the masked language modeling and next sentence prediction objective.

\textbf{\BERTL} is our implemented BERT, which can support input sequences with up to $1{,}024$ tokens. % Since training \BERTL from scratch would be time-consuming, 
We initialize the parameters of \BERTL with that of BERT, where the additional position embedding is initialized with the first $512$ ones.  After that, we continue to train it on Wikipedia\footnote{https://en.wikipedia.org/} for $22$k steps.

\textbf{DistilBERT}~\cite{DistilBERT} is the most popular distilled version of BERT, which leverages the knowledge distillation to learn knowledge from the BERT model. We use the $6$-layer DistilBERT released by Hugging Face\footnote{https://github.com/huggingface/transformers}.  In addition,
we use the same method to distill BERT with $3$ layers to obtain DistilBERT$_3$.

\textbf{DeFormer}~\cite{DeFormer} is designed for question answering, which encodes questions and passages separately in lower layers. It pre-computes all the passage representation and reuses them to speed up the inference. In our experiments, we do not count DeFormer's pre-computation.

\textbf{PoWER-BERT}~\cite{PoWER-BERT} is mainly designed for text classification, which also decreases the length of a sequence as layer increases. It adopts the Attention Strategy  to measure the significance of each token and always selects tokens with the highest attention. Given a length penalty, PoWER-BERT searchs a fixed length pruning configuration for all examples. %By adjust the length penalty of the soft-extract layer, PoWER-BERT provides various pruning intensity.

\textbf{DynaBERT}~\cite{DynaBERT} can not only  adjust model's width by varying the number of attention heads, but also provide an adaptive layer depth to satisfy different requirements. For a given speed demand, we report its best performance with all the  feasible width and depth combination options.

% \textbf{DynaBERT}

% \textbf{DeeBERT}~\cite{deebert} is a layer-wise pruning method designed for text classification. DeeBERT makes predictions in all layers and allows samples to exit earlier when their prediction confidence exceeds a given threshold. By adjusting the threshold, DeeBERT provides various pruning intensity.

\subsection{Datasets}
\begin{table*}[!t]
\resizebox{0.99\textwidth}{!}{
\centering
\begin{tabular}{l  r r r r r r r r r r r }
\toprule
\# Tokens  &   {SQuAD\,2.0} & {NewsQA}& {NaturalQA} & {RACE}  & {HotpotQA} & {TriviaQA} & {WikiHop}  &   {YELP.F}  & 20News. &  {IMDB}  & {Hyperp.} \\
\midrule
Average & 152 & 656 &248 &381 &1,988& 3,117& 1,499& 179& 551& 264 &755 \\
$95$th percentile &  275 & 878 &1,088& 542& 2,737& 4,004& 2,137& 498& 1,769& 679 &2,005 \\
\bottomrule
\end{tabular}
}
\caption{Average and $95$th percentile of context length of datasets in wordpieces. } %Hyper.: Hyperparisan. Yelp.F: Yelp full; Yelp.P: Yelp polarity; 
\label{tab:data_len}
% \vspace{-0.75em}
\end{table*}

\begin{table*}[!t]  
\resizebox{0.99\textwidth}{!}{
\centering
% \resizebox{\textwidth}{23mm}{
% \setlength{\tabcolsep}{7pt}
\begin{tabular}{l  c c c c c c c c c c c c c c c}
\toprule

% \multirow{2}{*}{Model}   & \multicolumn{2}{c}{SQuAD\,2.0} & \multicolumn{2}{c}{NewsQA} & \multicolumn{2}{c}{NaturalQA} &  \multicolumn{2}{c}{RACE} & \multicolumn{2}{c}{YELP.F}  &\multicolumn{2}{c}{IMDB}     \\

% & F1 & FLOPs & F1 & FLOPs & F1 & FLOPs & Acc. & FLOPs & Acc. & FLOPs & Acc. & FLOPs \\
% \midrule
% BERT &  77.12 &  1.00x  &  66.82&  1.00x & 78.32 & 1.00x  & 66.30 & 1.00x   & 69.93 &  1.00x   & 94.05& 1.00x \\
% DistilBERT$_6$ & 68.17& 2.00x  & 63.56 & 2.00x& 76.21 & 2.00x  & 52.63 & 2.00x & 69.42 & 2.00x  & 93.11 &2.00x    \\
% PoWER-BERT & -& -  & - &- & - &- & - &- & 68.21 & 2.90x &                     93.56 & 2.32x    \\
% DeFormer & 71.41 & 2.19x  & 60.68 & 2.01x & 74.34 & 2.34x & 64.27 &2.19x & - & - & - & -    \\ 
% DynaBERT & 74.42 & 2.00x  &   & 2.00x&   & 2.00x  &   & 2.00x &   & 2.00x  & 94.00 &2.00x    \\
% \midrule
% \bf{\DTBERTD} & 74.62 &  2.29x &65.09 & 2.36x & 79.03 & 2.51x & 64.33 & 2.19x &   69.76 & 2.65x  & 93.60  & 2.26x  \\
% \bf{\DTBERTS} & 71.75 &3.07x & 65.36 & 2.96x & 78.11 & 3.74x & 53.40 & 4.10x & 69.80 &  4.05x  & 92.64 & 4.13x \\
\multirow{2}{*}{Model}   & \multicolumn{2}{c}{SQuAD\,2.0} & \multicolumn{2}{c}{NewsQA} & \multicolumn{2}{c}{NaturalQA} &  \multicolumn{2}{c}{RACE} & \multicolumn{2}{c}{YELP.F} & \multicolumn{2}{c}{20NewsGroups} & \multicolumn{2}{c}{IMDB}     \\

& F1 & FLOPs & F1 & FLOPs & F1 & FLOPs & Acc. & FLOPs & Acc. & FLOPs & Acc. & FLOPs & Acc. & FLOPs \\
\midrule
BERT &  77.12 &  1.00x  &  66.82&  1.00x & 78.32 & 1.00x  & 66.30 & 1.00x   & 69.93 &  1.00x  & 86.66& 1.00x & 94.05& 1.00x \\
DistilBERT$_6$ & 68.17& 2.00x  & 63.56 & 2.00x& 76.21 & 2.00x  & 52.63 & 2.00x & 69.42 & 2.00x  & 85.85 & 2.00x&93.11 &2.00x    \\
PoWER-BERT & $-$& $-$  & $-$ &$-$ & $-$ &$-$ & $-$ &$-$ & 67.37 & 2.75x & 86.51 & 2.91x&                   92.02 & 3.05x    \\
DeFormer & 71.41 & 2.19x  & 60.68 & 2.01x & 74.34 & 2.34x & 64.27 &2.19x & - & $-$ & $-$ & $-$ & $-$ & $-$   \\ 
% DynaBERT & 74.83 & 2.00x  & 65.66  & 2.00x&  77.22 & 2.00x  &  59.85 & 2.00x & 69.84& 2.00x & 86.03 & 2.00x  & 94.00 &2.00x    \\
DynaBERT & 74.83 & 2.00x  & 66.13  & 2.00x&  78.14 & 2.00x  &  \textbf{65.38} & \textbf{2.00x} & 69.96& 2.00x & 86.03 & 2.00x  & \textbf{94.00} &\textbf{2.00x}    \\
\midrule
\bf{\DTBERTD} & \textbf{75.66} &  \textbf{2.08x} & \textbf{67.18} & \textbf{2.05x} & \textbf{79.03} & \textbf{2.51x} & \textbf{65.00} & \textbf{2.16x} &   69.97 & 2.19x  &  87.44& 4.22x&  \textbf{93.60}  & \textbf{2.26x}  \\
\bf{\DTBERTS} & 71.75 &3.07x & 65.36 & 2.96x & 78.11 & 3.74x & 53.40 & 4.10x & \textbf{70.04} &  \textbf{2.83x} & \textbf{86.58} & \textbf{5.90x} & {92.64} & 4.13x \\
\bottomrule
\end{tabular}
}
\caption{Comparison of performance and FLOPs (speedup) between \DTBERT and baselines. %We highlight our model in bold.
}
\label{tab:main_results}
% \vspace{-1em}
\end{table*}

To verify the effectiveness of reducing the sequence length, we evaluate \DTBERT on several tasks with relatively long context, including question answering and text classification. Table~\ref{tab:data_len} shows the context length of these datasets. We adopt seven question-answering datasets, including SQuAD 2.0~\cite{SQuAD2}, NewsQA~\cite{newsqa}, NaturalQA~\cite{naturalqa}, RACE~\cite{RACE}, HotpotQA~\cite{HotpotQA}, TriviaQA~\cite{Triviaqa} and WikiHop~\cite{WikiHop}. And we also evaluate models on four  text classification datasets, including YELP.F~\cite{YELP}, IMDB~\cite{IMDB},  20NewsGroups (20News.)~\cite{20NewsGroups}, and Hyperpartisan (Hyperp.)~\cite{Hyperpartisan}.  
Among them, HotpotQA, TriviaQA and WikiHop possess abundant contexts for reading, while the performance of question answering (QA) models heavily relys on the amount of text they read. To fairly compare BERT and \BERTL,  we split the context into slices and apply a shared-normalization training objective~\cite{DocQA} to produce a global answer candidate comparison across different slices for the former two extractive QA datasets. And we average the candidate scores in all slices for WikiHop.
Details of all datasets are shown in the Appendix.% due to the space limit. 

\subsection{Experimental Settings}

%For all datasets except Hyperpartisan, we adopt the data split in the original paper.
We adopt a maximum input sequence length of $384$ for SQuAD\,2.0, $1{,}024$ for long-text tasks and $512$ for others. We use the Adam optimizer~\cite{Adam} to train all models. The detailed training configuration is shown in the Appendix. 
% for a fair comparison

For the RL training, we sample $8$ action sequences each time and average their rewards as the reward baseline. In the second training process which aims to warm up the policy network, we employ $20\%$ imitation learning steps for question answering tasks and $50\%$ steps for text classification tasks. We search the number of token reduction module $T\in[1,2,3]$. And we find the models with $T=2$ gets similar quality and speed trade-offs as the models with $T=3$, and both of them perform better than models with $T=1$. Thus we adopt $T=2$ for simplification.  We denote the pruned models from BERT, \BERTL and DistilBERT$_6$ as \DTBERTD, \DTBERTL, \DTBERTS, respectively. 
For {BERT} and {\BERTL}, we attach the token reduction modules before the second and the sixth layers. For DistilBERT$_6$, we insert the token reduction modules before the second and the fourth layers.

To avoid the pseudo improvement by pruning padding for \DTBERT, we evaluate all models with input sequences without padding to the maximum length. For each dataset, we report the F1 scores or accuracy (Acc.), and the FLOPs speedup ratio compared to the BERT model. The model's FLOPs are consistent in the various operating environment. Therefore, it is convenient to estimate and compare the models' inference time by FLOPs.

%FLOPs accumulates the number of floating-point operations performed by a model for a single process. 
%Model's FLOPs on a given example is consistent in the various operating environment, thus it is convenient to estimate and compare the models' inference time by FLOPs.

\subsection{Overall Results}

\begin{figure}[!t]
%\begin{tabular}{cc}   

\begin{minipage}{0.49\linewidth}
  \centerline{\includegraphics[width= 1.15\textwidth]{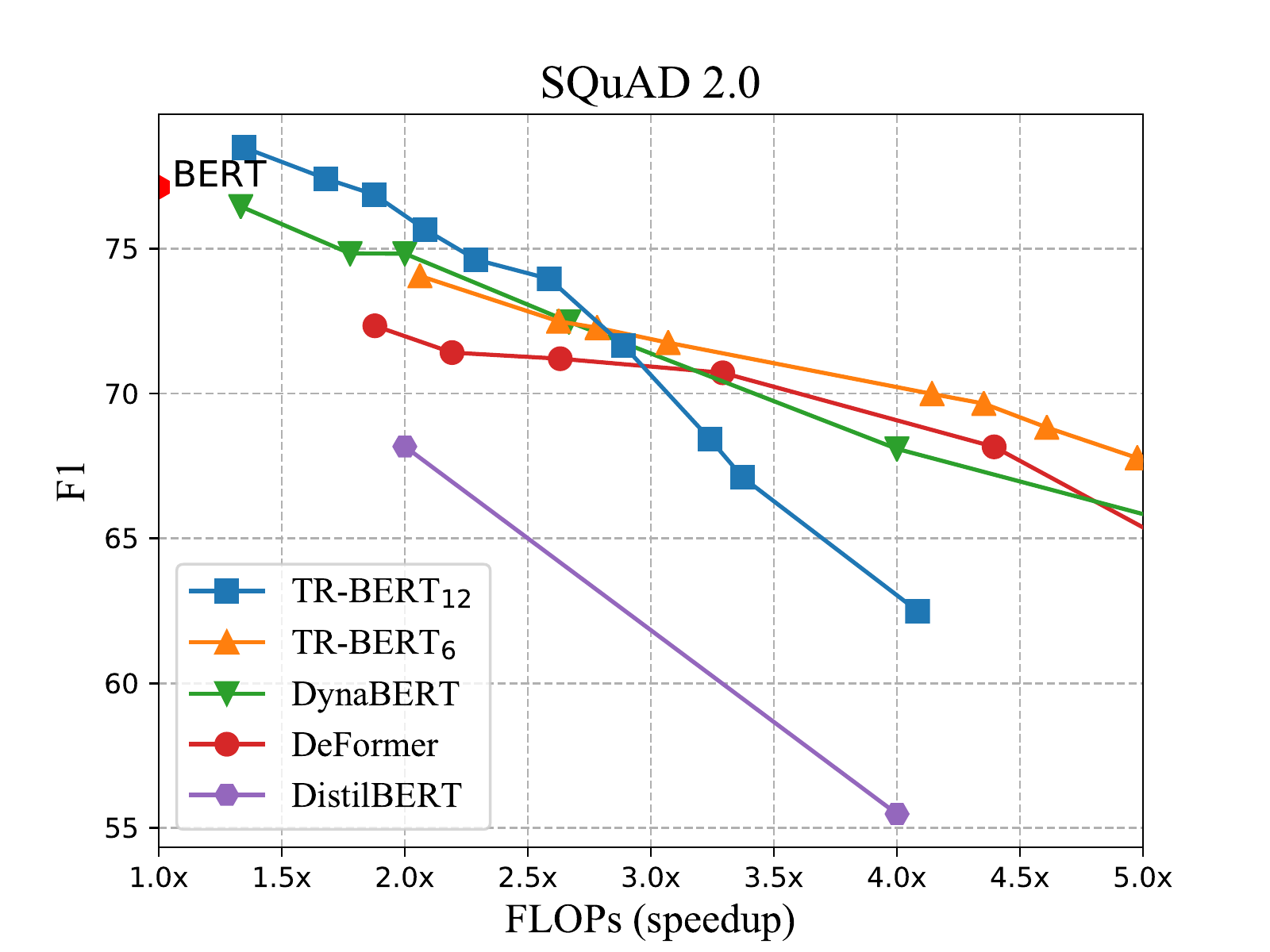}}
\end{minipage}
\hfill
\begin{minipage}{0.49\linewidth}
  \centerline{\includegraphics[width= 1.15\textwidth]{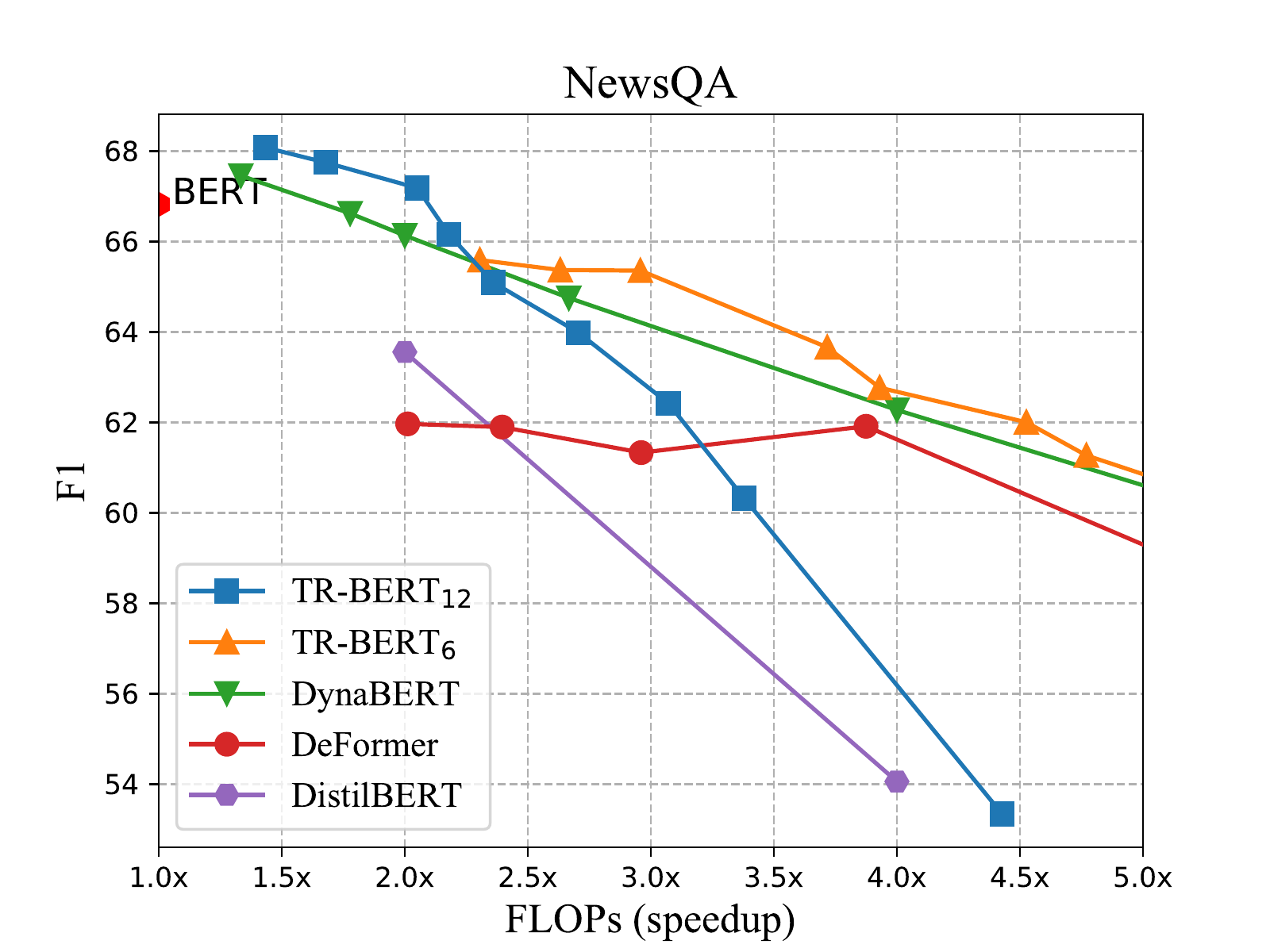}}
\end{minipage}
\hfill
\begin{minipage}{0.49\linewidth}
  \centerline{\includegraphics[width= 1.15\textwidth]{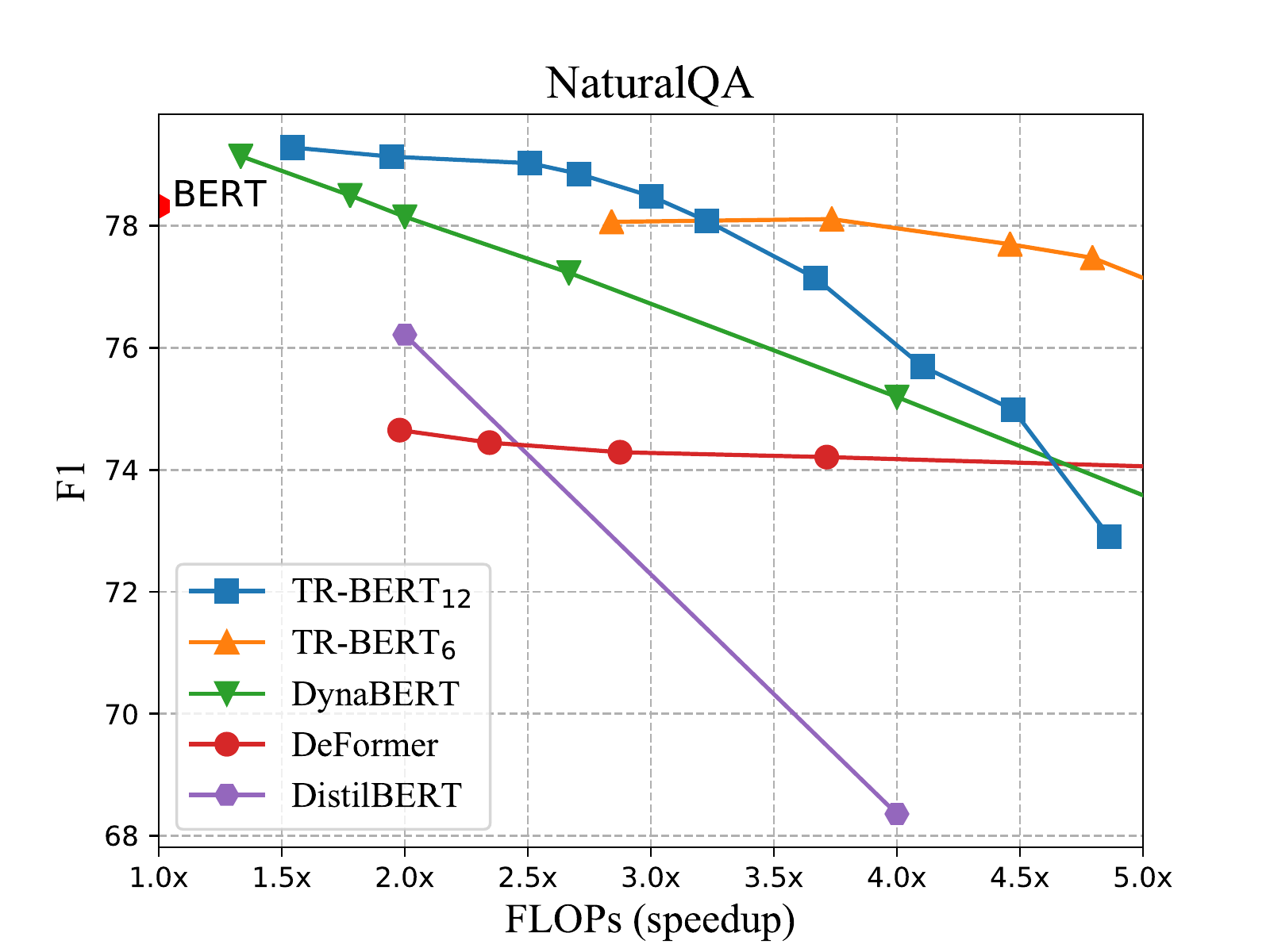}}
\end{minipage}
\hfill
\begin{minipage}{0.49\linewidth}
  \centerline{\includegraphics[width= 1.15\textwidth]{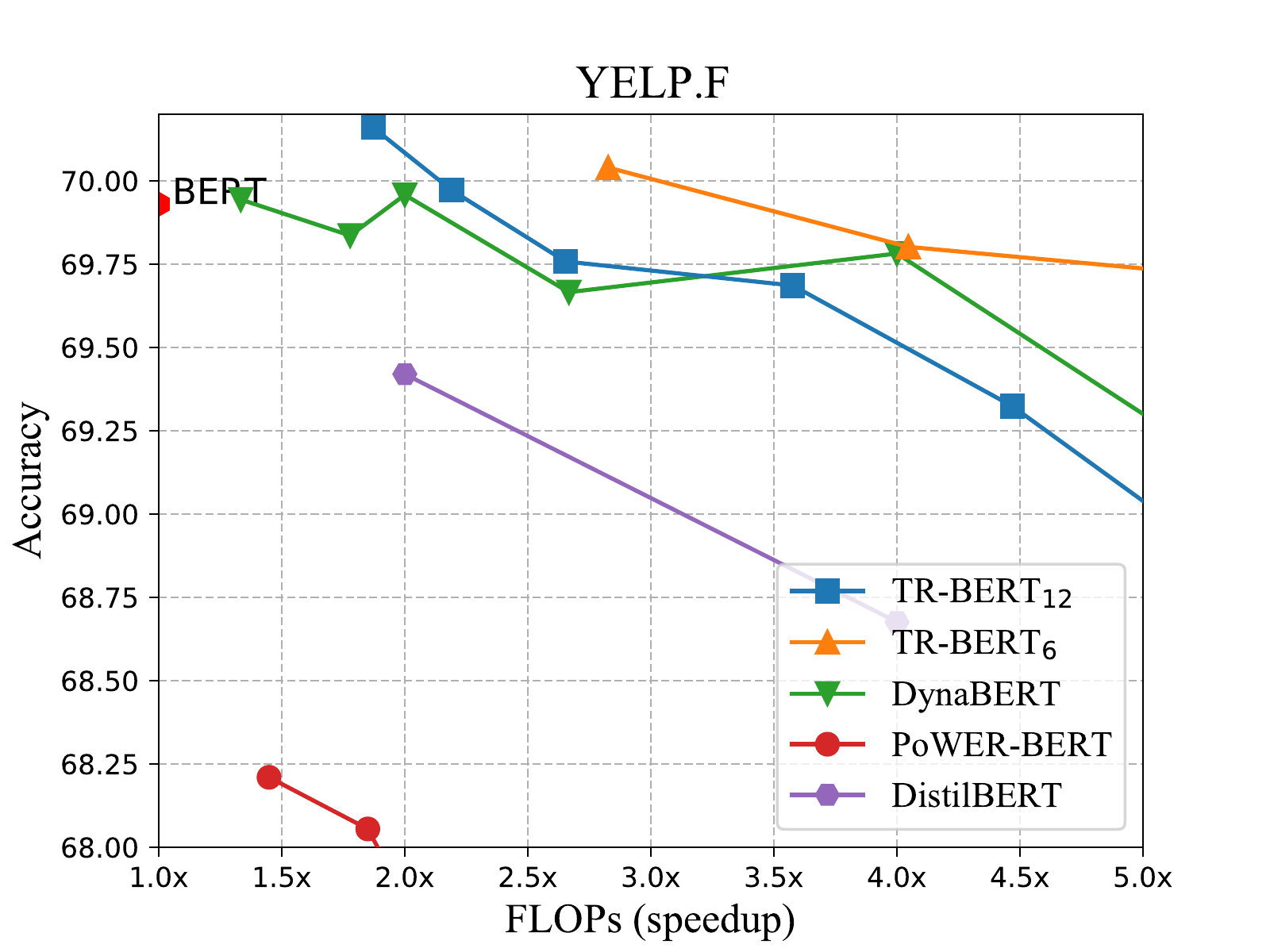}}
\end{minipage}

\caption{Quality and efficiency trade-offs for \DTBERTD and \DTBERTS. }
\label{fig:compare}
% \vspace{-1.5em}
\end{figure}

The comparison between \DTBERT and the baselines are shown in Table~\ref{tab:main_results} and Figure~\ref{fig:compare}. We adjust the length penalty coefficient of \DTBERT for an intuitional comparison. From the experimental results, we have the following observations:  

(1) \DTBERTD achieves higher performance while using less computation on all span-extraction QA datasets compared to all the baselines. For example, \DTBERTD outperforms DynaBERT by  $1.8$ F1 with faster speed. \DTBERTD even achieves better performance than BERT at low speedup rate, which demonstrates that discarding some redundant information in the top layer helps to find the correct answer.  For multiple-choice RACE, \DTBERTD achieves better performance than DeFormer while doesn't need to pre-compute the passage representation. %We observe that our  model can keep high prediction confidence when skip some important passage tokens for all the options. Hence, DTBERT's performance in multiple-choice QA is not as good as that of span-extraction QA.

(2) \DTBERTS  performs better than PoWER-BERT by a margin in text classification tasks. It shows that the fixed pruning configuration and the  attention-based selection strategy adopted by PoWER-BERT may not be flexible to accelerate inference for various input sequences. In contrast, our   dynamic token selection can automatically determine the proper pruning length and tokens for each example according to the actual situation, which leads to a more effective model acceleration.

%Attention Strategy for token selection may not be general for all tasks,  while our dynamic token selection keeps strong ability across all the evaluation tasks.

Overall, \DTBERT retains most of BERT's performance though it omits lots of token interactions in the top layers. It shows that \DTBERT learns a satisfactory token selection strategy through reinforcement learning, and could effectively reduce the redundant computation of tokens that have been extracted enough information in the bottom layers.

%has extracted enough information for the skipped tokens in the low layers. And reducing their redundant computation can significantly accelerate the prediction without sacrificing much performance. 

% (1) \DTBERT  can increase the inference speed of BERT by two times, while the performance drops of most datasets are tiny;
% (2) \DTBERT achieves better or comparable performance to layer-wise pruning  DistilBERT baseline on text classification tasks; 
% (3) Compared to DistilBERT, \DTBERT achieves higher performance while using less time in question answering tasks, which is generally more difficult than text classification task.

% BERT maintains a full-length sequence on all layers, which may lead to over-computation for unnecessary tokens. 
% From the experimental results, we find \DTBERT retains most of BERT's performance though it omits lots of token interactions in the high layers. It demonstrates that \DTBERT has extracted enough information for the skipped tokens in the low layers. And reducing their redundant computation can significantly accelerate the prediction without sacrificing much performance. 

% The last one shows the latency (speedup) on a RTX 2080 Ti GPU.
%  BERT and DistilBERT are viewed as special cases of DTBERTs with $1$x and $2$x speedup, respectively.

\subsection{Fuse Layer-wise and Token-wise Pruning}

Since layer-wise pruning and token-wise pruning are compatible, we also explore the incorporation of these two pruning strategies. We apply our dynamic token reduction on the $6$-layer DistilBERT to obtain \DTBERTS. The trade-off comparison of \DTBERTD and \DTBERTS is shown in Figure~\ref{fig:compare}, from which we have the following findings: 

(1) In general, as the speedup ratio increases, the performance of all models  decrease, which indicates that retaining more token information usually results in a more potent model.

%(2) \DTBERT beats BERT on three question answering tasks. It demonstrates that discard redundant information in high layer is beneficial for finding the proper answer; 

(2) \DTBERTS consistently outperforms \DTBERTD on all tasks at a high speedup ratio. In this situation, the budget doesn't allow enough tokens to go through the top layers. \DTBERTS makes a more elaborate pruning than \DTBERTD at bottom layers to obtain a better effectiveness.

(3) At low speedup ratio, \DTBERTD performs better than \DTBERTS on the question answering tasks, but worse on the text classification tasks. In general, a deep Transformer architecture can offer multi-turn feature extraction and information propagation, which can meet the complex reasoning requirements for question answering. In contrast, the result of text classification usually depends on the keywords in the context, for which a shallow model is an affordable solution. To obtain a better trade-off, we can flexibly employ a deep and narrow model for question answering and a shallow and wide model for text classification.%we are flexible to employ a deep and narrow model for question answering and a shallow and wide model for text classification. 

%We apply our dynamic token reduction on the distilled $6$-layer DistilBERT to obtain the DT-DistilBERT model. From Table~\ref{tab:main_results}, we found that DT-DistilBERT can further speed up the DistilBERT with minimal performance drop, which demonstrates our method's generalization.

\begin{table}[!t]
\small
\centering

\begin{tabular}{l  c c c c c c c c }
\toprule

\multirow{2}{*}{Model}   & \multicolumn{2}{c}{HotpotQA}  &  \multicolumn{2}{c}{TriviaQA}  \\

& F1 & FLOPs & F1 & FLOPs   \\
\midrule
BERT &  57.33 &  1.00x & 68.75& 1.00x\\
\BERTL & 65.45 & 0.91x & 69.69 & 0.92x \\
\bf{\DTBERTL} & \textbf{65.57} & \textbf{1.56x} & \textbf{70.41} & \textbf{1.20x} \\
\midrule

\multirow{2}{*}{Model}   & \multicolumn{2}{c}{WikiHop}  &\multicolumn{2}{c}{Hyperparisan}   \\
 & Acc. & FLOPs & Acc. & FLOPs   \\
 \midrule
BERT & 67.67 & 1.00x &  74.39 &  1.00x \\
\BERTL &  68.49 & 0.92x & 76.83 & 0.92x  \\
\bf{\DTBERTL}  & \textbf{67.78} & \textbf{4.65x} & \textbf{74.39} & \textbf{1.96x} \\

% \midrule

% \multirow{2}{*}{Model}   \\
%   \\
%  \midrule
% BERT \\
% \BERTL  \\
% \bf{\DTBERTL}  \\
\bottomrule
\end{tabular}

\caption{ Comparison performance and FLOPs (speedup) between \DTBERTL and BERTs with different maximum input sequence.}
% \vspace{-1.5em}
\label{tab:long_results}
\end{table}

\definecolor{Zeroselect}{RGB}{51,102,255}
\definecolor{Firstselect}{RGB}{0,0,255}
\definecolor{Secondselect}{RGB}{0,0,128}

\begin{table*}[h]
\small
\centering
\begin{tabular}{l | p {0.83\textwidth}}
\toprule
Dataset & Example \\
\midrule
\multirow{3}{*}{SQuAD 2.0}  &  Question: \,\,\,\,{\color{Secondselect}{How long did Western Europe control Cyprus?}} \\
& Paragraph: 
{\color{Secondselect} {The conquest}} 
{\color{Zeroselect} of} 
{\color{Secondselect} {Cyprus by the Anglo-Norman forces}} {\color{Zeroselect} of the} 
{\color{Secondselect} {Third Crusade}} 
{\color{Firstselect} {opened a new chapter in the history}} 
{\color{Zeroselect} of the} 
{\color{Secondselect} {island}}
{\color{Zeroselect}{,} }
{\color{Firstselect}{which would be}} 
{\color{Secondselect} {under Western European domination}} {\color{Zeroselect} for} 
{\color{Secondselect} {the following 380 years}}
{\color{Zeroselect}{.} }
{\color{Firstselect}{Although not part}}
{\color{Zeroselect} of} 
{\color{Firstselect}{a planned operation}} 
{\color{Zeroselect}, the} 
{\color{Secondselect} {conquest}}  
{\color{Firstselect}{had much more permanent results than initially expected}} 
{\color{Zeroselect}{.}}
\\
& Answer:  {\color{Secondselect} { 380 years}} \\
\midrule
IMDB Positive &  \textcolor{Zeroselect}{The}
 {\color{Firstselect} {Buddy Holly Story}} 
 {\color{Secondselect}  {is a great biography}} 
 {\color{Zeroselect} with a} 
 {\color{Firstselect} {super}} 
 {\color{Secondselect}{performance}}
 {\color{Zeroselect} from}
 {\color{Firstselect} {Gary Buse}}
 {\color{Zeroselect}.} 
{\color{Firstselect} {Busey did}} 
{\color{Zeroselect}{his own}}
{\color{Firstselect} {singing}} 
{\color{Zeroselect} for} 
{\color{Secondselect}{this}} 
{\color{Firstselect} {film}} 
{\color{Zeroselect} and he} 
{\color{Secondselect} {does}} 
{\color{Zeroselect} a}  
{\color{Secondselect} {great job}}
{\color{Zeroselect} .} \\
\midrule
IMDB \,Negative & \textcolor{Zeroselect}{I} {\color{Secondselect} {was really excited}}  {\color{Firstselect} {when} }{\color{Zeroselect}I} {\color{Firstselect} {read}} {\color{Zeroselect}``The Canterville} {\color{Firstselect} {Ghost}}{\color{Zeroselect}''} {\color{Firstselect} {would}}  {\color{Zeroselect}be} {\color{Firstselect} {shown}} {\color{Zeroselect}on} {\color{Firstselect} {TV}} {\color{Zeroselect}.} {\color{Secondselect} {However}} {\color{Zeroselect} , I} \textcolor{Secondselect} {was deeply disappointed} \textcolor{Zeroselect}{. I} \textcolor{Secondselect} {loved} {\color{Zeroselect}the} {\color{Firstselect} {original}} {\color{Zeroselect}story} {\color{Firstselect}{ {written}} {\color{Zeroselect}by} {\color{Firstselect} {Oscar Wilde}} {\color{Zeroselect}and}  {\color{Secondselect}  {sadly nothing}}  {\color{Zeroselect}of that} {\color{Firstselect} {was}}}  {\color{Secondselect} {transferred}} {\color{Zeroselect}by the} {\color{Secondselect}{movie}} {\color{Zeroselect}.}\\

\bottomrule
\end{tabular}
\caption{Case study of the token reduction. Skipped tokens in first token reduction module  are colored with {\color{Zeroselect} {light blue}}. Skipped tokens  in the second token reduction module are colored with {\color{Firstselect} {blue}}. The final selected token are colored with {\color{Secondselect} {dark blue}}. }
\label{tab:case_study}
% \vspace{-1.5em}
\end{table*}
 
% Pos.: Positive, Neg.: Negative. 

\subsection{Results on Long-text Tasks}

With token pruning, \DTBERT is able to process a longer sequence. %in order to extract information from long-range token interactions. 
We apply our dynamic token pruning strategy on \BERTL,  which can process sequence with up to $1{,}024$ tokens, to obtain   \DTBERTL , and conduct experiments on four datasets with longer documents, including HotpotQA, TriviaQA, WikiHop and Hyperparisan. Results on long-text tasks are shown in Table~\ref{tab:long_results},  from which we have the following observations:  

(1) \BERTL achieves better performance than BERT, especially on HotpotQA and WikiHop, which require the long-range multi-hop reasoning; 

(2) Compared to the vanilla BERT,   \DTBERTL achieves $8.2\%$ F1 improvement with $1.56$x speedup on HotpotQA, obtains $1.7\%$ F1 improvement with $1.24$x speedup on TriviaQA, gains $4.65$x speedup on WikiHop and $1.96$x speedup on Hyperparisan without performance drops.  Compared to BERT which can only deal with up to $512$ tokens at a time,  \BERTL considers a longer-range token interaction and obtains a more complete reasoning chain. However, the running time of \BERTL also increase as the input sequence's length extends, which poses a challenge to the utilization of longer text.  \DTBERTL inherits the broader view from \BERTL to get a better performance with a faster inference. Moreover, the inference acceleration effect of  \DTBERTL  is relatively better than \DTBERT within $512$ tokens, which is coincident to the above complexity analysis section.  With a longer sequence, \DTBERT can achieve extra speedup , because it significantly saves the time of the Self-Attention module, which demonstrates that \DTBERT can be further applied to process much longer tokens with limited computation.

\subsection{Case Study}
% 选词可视化

To investigate the characteristics of the selected tokens, we conduct a detailed case study on various datasets. As shown in Table~\ref{tab:case_study}, \DTBERT chooses to abandon the function word, such as \emph{the, and, with}, in the first token reduction module as the first module is placed at the bottom layer of BERT.  The second token reduction module is placed at the middle layer of BERT, and we could observe that it is used to retaining task-specific tokens. In the first example about question answering, the second token reduction module maintains the whole question and the question-related tokens from the context for further propagating messages.  In the second and third examples about movie review sentimental classification, the second token reduction module chooses to select sentimental words, such as \emph{great, excited, disappointed} to determine whether the given sequence is positive or negative.  

Although we train the token reduction module without direct human annotations, \DTBERT can remain the meaningful tokens in the bottom layer and select the higher layer's task-relevant tokens. It demonstrates that the pruned network's ground-truth probability is an effective signal to facilitate the reinforcement learning for token selection.

\section{Related Work}

Researchers have made various attempts to accelerate the inference of PLMs, such as quantization~\cite{Q-BERT,TernaryBERT}, attention head pruning~\cite{16heads, DynaBERT}, dimension reduction~\cite{mobileBERT, BertLottery}, and layer reduction~\cite{DistilBERT, BERT-PKD, TinyBERT}. In current studies, one of the mainstream methods is to dynamically select the layer number of Transformer layers to make a on-demand lighter model~\cite{LayerDrop, deebert, FastBERT}. However,  these methods operate at the whole text and they cannot perform pruning operations in a smaller granularity, such as the token-level granularity.
%For example, DistilBERT~\cite{DistilBERT}, PKD-BERT~\cite{BERT-PKD},  and Tiny-BERT~\cite{TinyBERT} leverage knowledge distillation to learn a shallow student model from a deep teacher model.
%LayerDrop~\cite{LayerDrop} employs a dropout function on Transformer layers in the training process, which allows the model to select an on-demand sub-network during the inference period. DeeBERT~\cite{deebert} and FastBERT~\cite{FastBERT}, designed for text classification, measures the uncertainty of each layer to decide whether to exit dynamically. However, all these methods operate at the whole text and cannot perform pruning operations in a smaller granularity. 

To consider the deficiencies of layer-level pruning methods, researchers decide to seek solutions from a more meticulous perspective by developing methods to extend or accelerate the self-attention mechanism of the Transformer. For example, Sparse Trasformer~\cite{SparseTrasformer}, LongFormer~\cite{longformer} and Big Bird~\cite{bigbird} employ the sparse attention  to allow model to handle long sequences. However, these methods only reduce the CUDA memory but cannot be not faster than the full attention. %Reformer~\cite{Reformer} introduces locality sensitive hashing to group elements and conduct attention mechanism within each group. 
%And Linformer~\cite{Linformer} uses singular value decomposition to approximate the attention matrix into a low-rank matrix for linear computation.
%then computes with compressed sequence 
Besides, researchers also explore the feasibility of reducing the number of involved tokens. For example, Funnel-Transformer~\cite{Funnel-Transformer}  reduces the sequence length with pooling for less computation, and finally up-samples it to the full-length representation.
Universal Transformer~\cite{UniversalTransformers}  builds a self-attentive recurrent sequence model, where each token uses a dynamic halting layer. And DynSAN~\cite{DynSAN} applies a gate mechanism to measure the importance of tokens for selection. % and multiply the gate value to the residual of the Transformer for training. 
Spurred by these attempts and positive results, we introduce \DTBERT in this study, which can creatively prune the network at the token level. To be specific, our work aims to accelerate the Transformer by deleting tokens gradually as the layer gets deeper. Compared with these models, \DTBERT is easy to adapt to the current PLMs models without a significant amount of pre-training and is flexible to adjust the model speed according to different performance requirements.

The main idea of \DTBERT is to select essential elements and infuse more computation on  them, which is widely adopted in various NLP tasks. ID-LSTM~\cite{ID-LSTM} selects important and task-relevant words to build sentence representation for text classification. SR-MRS~\cite{SR-MRS} retrieves the question-related sentences to reduce the size of reading materials for question answering. \DTBERT can be viewed as a unified framework on the Transformer for the important element selection, which can be easy to be applied in wide-range tasks.

\section{Conclusion and Future Work}
In this paper, we propose a novel method for accelerating BERT inference, called \DTBERT, which prunes BERT at token-level granularity. Specifically, \DTBERT utilizes reinforcement learning to learn a  token selection policy, which is able to select general meaningful tokens in the bottom layers and select task-relevant tokens in the top layers. Experiments on eleven NLP tasks demonstrate the effectiveness of \DTBERT as it accelerates BERT inference by $2$-$5$ times for various performance demand. Besides, \DTBERT achieves a better quality and speed trade-off on long-text tasks, which shows its potential to process large amounts of information in the real-world applications. % As a token-wise pruning method, \DTBERT is also compatible with the layer-wise methods to make a more reasonable model structure, e.g., narrow and deep models for question answering.

In the future, we would like to attempting to apply \DTBERT in the pre-training process of PLMs. Through the automatically learned token reduction module, it is possible to reveal how BERT stores syntactic and semantic information in various tokens and different layers. And it's also worth speeding up the time-consuming pre-training process.  

%(2) To train the policy network, we inherit the parameters of the backbone network from a fine-tuned model, which, nevertheless, is costly trained with full-length tokens across layers. Hence, it is worth developing  DT-BERT's training from scratch to speed up the convergence. 

\section*{Acknowledgement}
This research is mainly supported by Science \& Tech Innovation 2030 Major Project "New Generation AI" (Grant no. 2020AAA0106500) as well as supported in part by a grant from the Institute for Guo Qiang, Tsinghua University.

\bibliography{anthology,custom}
\bibliographystyle{acl_natbib}

\appendix
\label{sec:appendix}

\section{Details of Datasets}

We evaluate models on seven question-answering datasets, including SQuAD\,2.0~\cite{SQuAD2}, NewsQA~\cite{newsqa}, NaturalQA~\cite{naturalqa}, RACE~\cite{RACE}, HotpotQA~\cite{HotpotQA}, TriviaQA~\cite{Triviaqa} and WikiHop~\cite{WikiHop}. Besides, we evaluate models on four long-text classification datasets, including YELP.F~\cite{YELP}, 20NewsGroups~\cite{20NewsGroups}, IMDB~\cite{IMDB}, and Hyperpartisan~\cite{Hyperpartisan}. We use the MRQA~\cite{MRQA} version of NewsQA and NaturalQA. Details of all evaluation datasets are shown below:

 \textbf{SQuAD\,2.0}~\cite{SQuAD2} is a large-scale reading comprehension dataset. Compared to its former SQuAD\,1.0~\cite{SQuAD}, SQuAD\,2.0 involves  $54$k unanswerable questions,  which empirically makes the task much harder. For SQuAD\,2.0, \DTBERT not only needs to find the question-relevant tokens, but also requires to check sufficient evidence to make a waiver decision when no answer is available. To predict the answer, we attach a span predictor on the top of BERTs and  set the answer of the  unanswerable question as a span of $[0,0]$. 

 \textbf{MRQA}~\cite{MRQA} integrates several existing datasets to a unified format, which provides a single context within $800$ tokens for each question,   ensuring at least one answer could be  accurately found in the context. We adopt the NewsQA and NaturalQA dataset from the MRQA benchmark.
%consists of over $100$k question-answer pairs on $10$k news articles from CNN. We use NewsQA of MRQA~\cite{MRQA} version, which 

  \textbf{RACE}~\cite{RACE} is composed of $98$k multiple-choice questions collected from English examinations. The model needs to figure out the correct answer from four options for a given question and passage. Passages in RACE cover a variety of topics, which can examine the generalization of our token selection.

  \textbf{HotpotQA}~\cite{HotpotQA} is an extractive question answering dataset, which  requires multi-hop reasoning over multiple supporting documents for answering $113$k questions. We adopt the full-wiki setting for HotpotQA, which requires models to find answers from a large-scale corpus.   We employ the retriever and re-ranker in Transformer-XH~\cite{Transformer-XH} to obtain question-related paragraphs and merge them into one document. In HotpotQA, models are required to reason over bridge entities or check multiple properties in different positions,  which brings challenges to the token selection of \DTBERT in considering the global information. We concatenate two positive paragraphs and several negative paragraphs to make the reading material for training, which contains up to $1{,}024$ tokens. And we concatenate the re-rank passages in order for evaluating. For the vanilla BERT, we apply a shared-normalization training objective~\cite{DocQA} to produce a global answer candidate comparison across two parts of the document.
%$63\%$ of the answers need to be inferred through either bridge entities or checking multiple properties in different positions, which brings challenges to the token selection of \DTBERT. 

  \textbf{TriviaQA}~\cite{Triviaqa} has more than $95$k question-answer pairs authored by Trivia enthusiasts. We use the Wikipedia setting of TriviaQA, which provides question-retrieved paragraphs from Wikipedia. We use the linear passage re-ranker in DocQA~\cite{DocQA} to re-rank these retrieved paragraphs and finally concatenate the first $1{,}024$ tokens as a new reading document. We also employ the shared-normalization training objective~\cite{DocQA} for the vanilla BERT.

% $512$ for BERT and $1024$ for BERT$_{1024}$. 

% The setting requires model to find answer from the whole Wikipedia, which is more challenging and match the actual requirement in industry.

  \textbf{WikiHop}~\cite{WikiHop} consists of $51$k questions, candidate answers, and supporting paragraphs triples.  It requires models to find multi-hop reasoning chains for choosing the correct answer. Due to the long length of the concatenation of supporting paragraphs, we follow the processing tactic in Longformer~\cite{longformer}, which splits the document into several parts and then averages their candidate scores. 

% Based on supporting contexts, model aims to choose correct answer candidate for the given question. 
% The dataset is designed for multi-hop reasoning, which also needs understanding across multiple documents.

  \textbf{YELP.F}~\cite{YELP} contains $1,569$k samples with review texts, which are obtained from the YELP Dataset Challenge in 2015. Yelp Review Full (YELP.F)  contains  five star classes.

\textbf{20NewsGroups}~\cite{20NewsGroups} comprises around $18$k newsgroups posts on $20$ topics. 

  \textbf{IMDB}~\cite{IMDB} consists of $50$k informal movie reviews from the Internet Movie Database. Each review is annotated as \emph{positive} or \emph{negative}.

  \textbf{Hyperpartisan}~\cite{Hyperpartisan} aims to decide whether a news article text follows a hyperpartisan argumentation. Hyperparisan only contains $645$ documents, which makes it a good testbed in a low-resource scenario. For the vanilla BERT, we adopt the max-pooling results of BERT sliding window. We split the data into training / validation / test set with a ratio of $9$:$1$:$1$, run each model five times, and report the median performance.

\section{Training Configuration}
We follow the configuration from previous work~\cite{BERT,longformer,FastBERT} for fine-tuning BERT, DistilBERT and \BERTL. Hyperparameters are shown in Table~\ref{tab:Hyperparameters}.

\begin{table}[h]
\centering
\resizebox{0.99\linewidth}{!}{
\begin{tabular}{l  r r r r r} 
\toprule
Dataset & Epoch & LR & WP & BSZ & Optimizer\\
\midrule

SQuAD & $2$ & $3e$-$5$ & $10\%$ & $32$ & Adam\\
NewsQA & $5$ & $3e$-$5$ & $10\%$ & $32$ & Adam\\
NaturalQA & $5$ & $3e$-$5$ & $10\%$ & $32$ & Adam\\
RACE & $5$ & $5e$-$5$ & $10\%$ & $32$ & Adam\\
HotpotQA & $6$ & $5e$-$5$ & $7.3\%$ & $32$ & Adam\\ 
TriviaQA & $5$ & $3e$-$5$ & $10.6\%$ & $32$ & Adam\\ 
WikiHop & $15$ & $3e$-$5$ & $1\%$ & $32$ & Adam\\ 
YELP.F & $3$ & $3e$-$5$ & $10\%$ & $32$ & Adam\\ 
20NewsGroups & $5$ & $3e$-$5$ & $10\%$ & $32$ & Adam\\ 
IMDB & $5$ & $3e$-$5$ & $10\%$ & $32$ & Adam\\ 
Hyperparisan & $15$ & $3e$-$5$ & 10\% & $32$ & Adam\\
\bottomrule
\end{tabular}
}
\caption{Hyperparameters of all the models in different datasets. LR: Learning rate;  BSZ: Batch size;  WP: Warmup proportion.   }
\label{tab:Hyperparameters}
\end{table}
To train \DTBERT, we first initialize \DTBERT with corresponding fine-tuned models, which are trained with a task-specific objective for $N$ epochs. After that, we maintain the same learning rate, warmup proportion and batch size for the latter two-step training: (1)  Freeze all the parameters except that of the policy network and conduct reinforcement learning to update the policy network for $\lceil {(N+1)/2} \rceil $ epochs; (2) Unfreeze all parameters and  train the entire network with the task-specific knowledge distillation objective and the reinforcement learning  objective simultaneously for  $N$ epochs.

\section{Actual Wall Time}
In practical applications, the wall time acceleration of TR-BERT is similar to the FLOPs acceleration. We evaluate our model on a single V100 GPU with $32$ batch size  on SQuAD. TR-BERT in Table~\ref{tab:main_results} with $2.08$x FLOPs speedup achieves $2.01$x actual inference time speedup.

\end{document}